\documentclass[conference,comsoc]{IEEEtran}
\usepackage[english]{babel}
\usepackage{graphicx}
\usepackage{subfigure}
\ifCLASSOPTIONcompsoc
  \usepackage[nocompress]{cite}
\else
  \usepackage{cite}
\fi
\usepackage{url}

\addto\captionsenglish{\renewcommand{\figurename}{Fig.}}
\addto\captionsenglish{\renewcommand*{\tablename}{TABLE}}

\begin{document}

\title{Multi-label Class-imbalanced Action Recognition in Hockey Videos via 3D Convolutional Neural Networks}

\author{
    \IEEEauthorblockN{Konstantin Sozykin, Stanislav Protasov and Adil Khan}
	\IEEEauthorblockA{Institute of Robotics, \\
	Innopolis University, Russia \\
	Email: \small\texttt{\{k.sozykin,s.protasov,a.khan\}@innopolis.ru}}
	\and
    \IEEEauthorblockN{Rasheed Hussain, Jooyoung Lee}
	\IEEEauthorblockA{Institute of Information Systems, \\
	Innopolis University, Russia \\
	Email: \small\texttt{\{r.hussain,j.lee\}@innopolis.ru}}
}

\maketitle

\begin{abstract}
Automatic analysis of the video is one of most complex problems in the fields of computer vision and machine learning. A significant part of this research deals with (human) activity recognition (HAR) since humans, and the activities that they perform, generate most of the video semantics. Video-based HAR has applications in various domains, but one of the most important and challenging is HAR in sports videos. Some of the major issues include high inter- and intra-class variations, large class imbalance, the presence of both group actions and single player actions, and recognizing simultaneous actions, i.e., the multi-label learning problem. Keeping in mind these challenges and the recent success of CNNs in solving various computer vision problems, in this work, we implement a 3D CNN based multi-label deep HAR system for multi-label class-imbalanced action recognition in hockey videos. We test our system for two different scenarios: an ensemble of $k$ binary networks vs. a single $k$-output network, on a publicly available dataset. We also compare our results with the system that was originally designed for the chosen dataset. Experimental results show that the proposed approach performs better than the existing solution.
\end{abstract}


\section{Introduction}
Automatic recognition of human activities is an exciting and challenging research area and it can has applications in fields, such as robotics\cite{Zhu:2009:HDA:1703775.1704035}, healthcare \cite{DBLP:journals/mbec/KhanLLK10,Lee_EMBC_2010}, sport analytics\cite{DBLP:journals/tmm/HuangSC06}, and security etc. G. Johansson \cite{Johansson1973} pioneered this area by developing the first method for modeling and analysis of human locomotion in visual data. Since then a significant amount of work has been done in this regard\cite{Aggarwal:2011:HAA:1922649.1922653}. 



\par Variations in motion or movement patterns may result because the same activity may be performed differently by different individuals as well as by the same person \cite{Bulling:2014:THA:2578702.2499621,Saputri_IJDSN_2014}. There are many reasons for this. For example, stress, time of the day, health and emotional states. From machine learning point of view, it is called high intraclass variability or variance problem.

\par On the other hand, there is interclass similarity; it is the case when two or more different classes have similar characteristics, but they are fundamentally different. Good and straightforward examples of this case are activities like walking and running (jogging). They have a higher visual similarity, but they are definitely from different action categories. 


\par The class imbalance  is the case when the classes are not represented equally. This could lead to a problem since many machine learning approaches (especially complex algorithms like neural networks) work well only if the number of observations for all classes are roughly equal. There are a number of methods in machine learning literature that can be used to handle this problem. For example, balancing the training data by means of oversampling or under-sampling, and class weight adjustment \cite{He:2009:LID:1591901.1592322}.

\par Furthermore, in real-life, it is common to have situations when at any given moment more than one action may happen. It happens because in the case of videos often multiple persons are present, and they may simultaneously interact with each other or with different objects. 
From machine learning point of view, it means that an observation may belong to multiple classes. Therefore, human action recognition problem may lead to the multi-label learning problem. Multi-label learning is a generalization of supervised learning with the assumption that observed instances can belong to more than one class simultaneously. As a large field of research, it has its own issues and associated methods. For example, it can require special loss functions or algorithms to work in $k$-output mode. More details can be found in \cite{DBLP:journals/tkde/ZhangZ14}.

\par Last but not the least, the domain of the chosen activities, such as home activities or sports activities, can further add to the complexity of the recognition task. In the previous paragraphs, we discussed the class imbalance and multi-label learning problems. In the case of sports action recognition from video,  they are both strongly present. For example, in many active team sports, such as hockey or soccer, \textit{Goal} is a very rare action compared to \textit{Running}. Therefore, even if we have many hours of video data it is difficult to collect enough samples of the \textit{Goal} class. On the other hand, in team sports, players may perform different actions at the same time. 


\par We chose neural networks, especially deep networks, for building our recognition system, since they offer real advantages. Firstly, deep learning and convolution neural networks (CNNs) have recently shown excellent performance in different complicated visual tasks. Examples of such visual tasks include but are not limited to image recognition \cite{DBLP:journals/corr/SimonyanZ14a}, object detection and recognition \cite{DBLP:journals/corr/RenHG015}, object tracking \cite{DBLP:journals/corr/ValmadreBHVT17} etc.
\par Next, with convolution-based feature extraction, we can learn not only the classification models but also the class representations \cite{Goodfellowbook}. By learning representations, we mean learning a set of abstract features that can efficiently represent each class. In general, if we have a better representation of some data, especially visual, we can do a better learning for related or similar tasks using these representations. 

\par We implement, test and compare two deep approaches for multi-label activity learning having class imbalance problem in hockey videos. The two approaches are: (i) An ensemble of $k$ binary networks, and (ii) A single multi-label $k$-output network. Also, we compare the results with a state-of-the-art existing solution \cite{DBLP:conf/iscas/CarbonneauRGG15}. 
\par We make 3D CNN our baseline and provide $F_{1}$ measure scores for a publicity available dataset \cite{DBLP:conf/iscas/CarbonneauRGG15}. We do it for 11 activities, instead of just three activities as was done in the previous work \cite{DBLP:conf/iscas/CarbonneauRGG15}. It will be useful for any researcher who is working on the same or similar problems. We believe that there are lots of areas where this dataset, with provided $F_{1}$ baseline scores, will be helpful. 

\section{Related Work}


\subsection{Traditional Approaches}
\par Most of the existing works on action recognition in sports video are based on traditional machine learning and computer vision methods. For example, \cite{DBLP:journals/corr/WaltnerMB14} proposed a method for learning and recognizing activities in a volleyball game. The authors concentrated on single player activity recognition and got 77.8 \% recognition accuracy. Their main idea was to build a context descriptors based on Histogram of Oriented Gradients (HOG), and Histogram of Optical Flow (HOF) features and employ Support Vector Machines (SVM) and Gaussian Mixture Models (GMM) as classifiers, for seven classes in six video in public datasets. 


\par In \cite{Tora_2017_CVPR_Workshops}, Tora et al. proposed a puck possession action recognition method of the hockey game. Their approach was based on aggregation of individual and context information with pre-trained CNN and further LSTM training. The system was evaluated using a dataset,which was obtained from SportLogiq and consisted of up to 5 events , and an average recognition accuracy of up to 49.2\% was achieved. 

\par For action recognition of the hockey game, Carbonneau et al. \cite{DBLP:conf/iscas/CarbonneauRGG15} presented a solution for play-break detection using STIP\cite{BMVC.23.124} detectors and SVMs. Although their system achieved a good performance, up to 90 \% recognition accuracy, their analysis was limited to only three activities. There are no baseline recognition scores for nine other action classes that are present in their dataset. 

\subsection{Deep Approaches}
\par Nowadays deep learning has shown excellent performance, especially in visual tasks, such as object recognition (\cite{DBLP:journals/corr/RenHG015}, image classification\cite{NIPS2012_4824}, and sports action recognition. For example, \cite{DBLP:journals/corr/IbrahimMDVM15} presents a CNN and Long Short-term Memory (LSTM) based architecture for learning hierarchical group activities in volleyball video dataset, which was collected from YouTube. The key idea in their work was to use fine-tuned AlexNet features (fc7 layer) as input to a two-staged LSTM classifier for person and group activity recognition. The approach yielded a recognition accuracy of 63 - 86\% for six activities.
     
\par In \cite{Karpathy:2014:LVC:2679600.2680211}, Karpathy et al. presented a Sport-M1 dataset collected by Standford Vision Lab, and multi-resolution CNN architecture that achieved 41.3 - 64.1\% average accuracy. The dataset was about various sports, and consisted of  487 activities.

\par Recently, Kay et al. from Google DeepMind team presented the Kinetics \cite{DBLP:journals/corr/KayCSZHVVGBNSZ17} dataset. It is a large-scale publicly available Youtube-based dataset that includes various sets of human actives, approximately over 400 activities within 300,000 videos. In their work three deep baseline approaches were presented, including 3D CNN, 2-stream CNN (with RGB and optical flow inputs), and CNN+LSTM models with performance in the range of  56 - 79 \% on the presented new dataset. 
 

\section{Methodology}

\par Let us first define the multi-label learning problem in the context of action recognition in hockey videos. Let $D=\left \{  (\mathbf{x}_i,\mathbf{y}_i)| 1\leq i \leq m \right \}$ be the multi-label training data. For the $i$-th multi-label instance $(\mathbf{x}_i,\mathbf{y}_i)$, $\mathbf{x}_i$ is a $d$-dimensional feature vector $(x_{i1}, x_{i2}, ...,x_{id})$ of real values, and $\mathbf{y}_i$ is the associated $k$-dimensional label vector $(y_{i1}, y_{i2}, ...,y_{ik})$ of binary values for $k$ possible classes (actions). For an unseen instance $\mathbf{x}$, the classifier $h(.)$ predicts $(y_{1}, y_{2}, ...,y_{k})$ as a vector of labels for $\mathbf{x}$.


\par One way to automatically extract features from video data is to apply CNNs, with typical convolution and pooling layers. In our work, we use 3D convolution and 3D pooling, which is a generalization of CNN operations, to perform feature extraction not only from a single image but also from a slice of frames 


\par As for how all of this is implemented as an end-to-end system, we implement and test two different strategies. First is an ensemble of $k$ independent single-label learning networks. It is a simple general idea for multi-label learning, where we split $k$ multi-label problem into $k$ binary learning problems, training  and evaluating $k$ classifiers independently for each of the $k$ classes. In literature, this method often is called binary relevance \cite{DBLP:journals/tkde/ZhangZ14}. 
To get multi-label prediction we just concatenate individual predictions into one vector. The second is a single multi-label $k$-output network. 


\par
\begin{figure}[ht]
	\centering
	\includegraphics[scale=0.3]{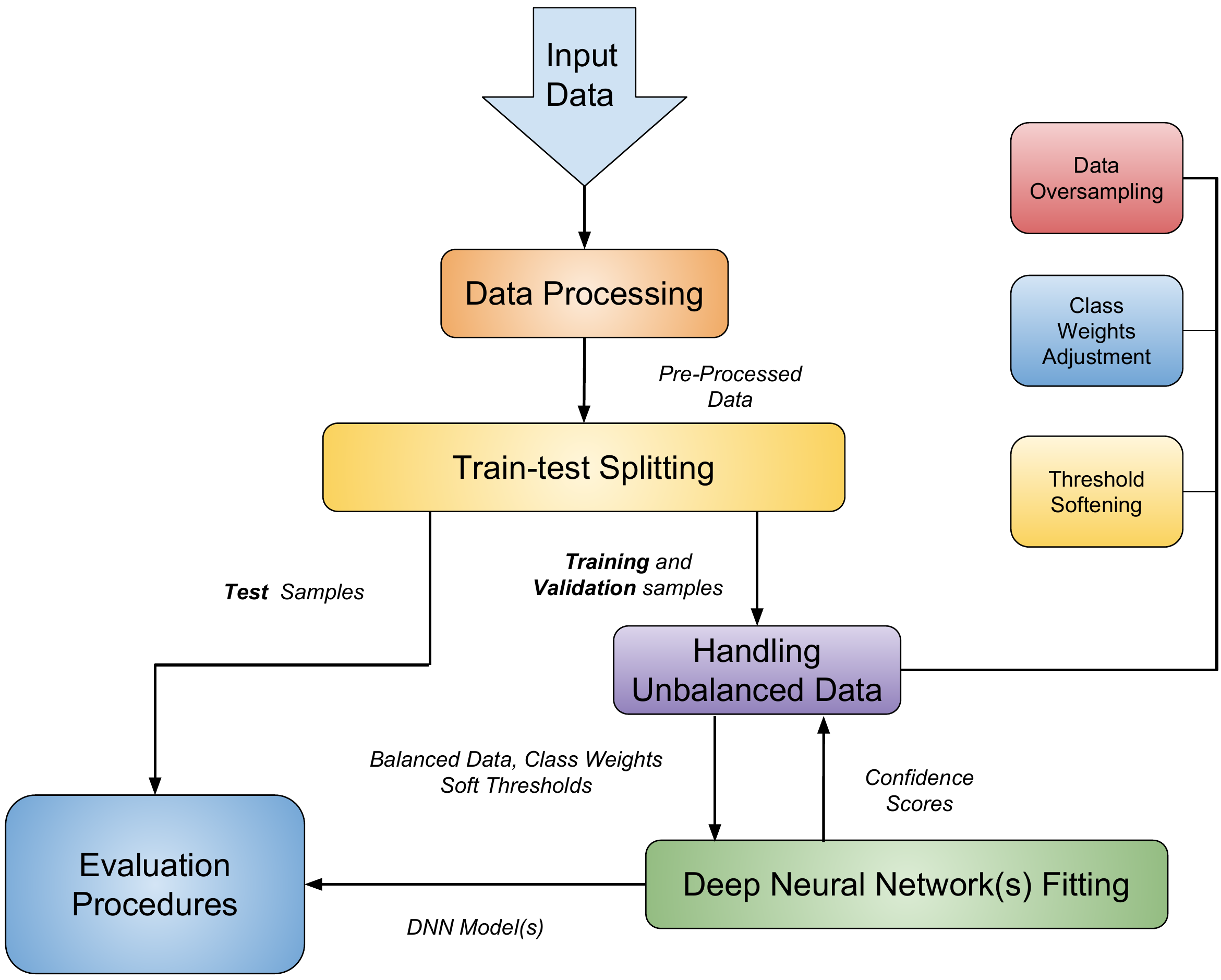}
	\caption{Graphical illustration of the research methodology that we followed in this work.}
    \label{work-approach-summary}
\end{figure}

 
We provide a graphical illustration of how the whole system is implemented and evaluated in Fig. \ref{work-approach-summary}.

\subsection{Data Preprocessing}
\subsubsection{Resizing} 
Main reason for resizing is the hardware limitations, since even a small batch of video data may require a lot of memory for processing. We resize each frame of a video. We found empirically that resizing by four times is optimal for our case. 

\par \textit{Data normalization}. This is a necessary step in neural network training since without normalization the loss gradient values could go unproportionate and could negatively affect the training process. In this work we perform standard normalization.

\par \textit{Windowing}. This means splitting the data, both the instances and their associated labels, into fixed-size units/sequences. We use overlapping sliding window protocol with a window size of 15, and an overlap of five frames. Doing so helps us in producing more samples. 

\par \textit{Sequence Labeling}. Each sequence of frames, produced in the previous step, must be associated with a single label vector. We apply the majority rule over each element of the associated 15 label vectors (for 15 frames) to produce the final label vector.

\par \textit{Training-test Splitting}. After the data preprocessing, we divide the entire data into two parts: training and test datasets, using a 70:30 split. Using the same split, we further divide the training data into training and validation datasets, which are used for training the models and selecting the appropriate values of the hyper-parameters. The test dataset is used in the end to evaluate the learned models.

\par \textit{Handling of Unbalanced data}. By this, we mean handling the class imbalance problem. In the case of the ensemble of $k$ binary networks, we apply a simple technique called oversampling \cite{DBLP:books/sp/datamining2005/Chawla05}. In this approach, when training each of the $k$ networks, we achieve balance by randomly adding copies of instances of the under-represented class.
However, oversampling is not an optimal method to use for solving the class imbalance problem in the case of single multi-label $k$-output network. The reason: an instance, in this case, may be associated with multiple labels, and randomly adding copies of such instances may affect the correlation among different labels. Therefore, for this case, we implement the following two-staged approach based on the concepts that are described in \cite{DBLP:journals/tkde/ZhangZ14}.
\par At the first stage, we use a technique called class weight adjustment, where the weight of a class is determined as 
\begin{equation}
w_i=log(\mu \frac{m}{m_i})
\label{eq:weight}
\end{equation}
where $w_i$ is the weight of the $i$-th class, $m$ is the total number of instances in the training dataset, $m_i$ is the number of instances that are associated with the $i$-th class, and $\mu$ is some constant in the range of $0\cdots1$. In our case, we set its value to 0.7, which is found empirically. It should also be noted that if (\ref{eq:weight}) returns a weight that is less than one, its value is set back to one. Thus the minimum possible weight of a class is one.   

\par At the second stage, we perform threshold softening for under-represented classes. By threshold, we mean the value against which the real-valued model output is going to be calibrated. To do this, we perform an initial training of the model using the calculated class weights, and a $k$-dimensional threshold vector whose elements are assigned to the default threshold $\alpha = 0.5$. After the initial training, the model is tested on the validation dataset to obtain the confidence scores (real-valued model output) for each class over all instances. The new threshold for the $i$-th class is then computed as
\begin{equation}
th_i=\alpha\frac{1}{w_i}c_i
\end{equation}
where $th_i$ is the new threshold, $\alpha$ is the default threshold, $w_i$ is the class weight, and $c_i$ is the maximum confidence score obtained for the class over all instances during the validation step. 
\par To conclude, in the case of single multi-label $k$-output network, class imbalance problem is resolved by assigning higher class weights to under-represented classes and using softer thresholds for the same.

\subsection{Network structure and Training Settings}

\par Our network is inspired by classical CNN architectures like ALEXNET\cite{NIPS2012_4824} and VGG\cite{DBLP:journals/corr/SimonyanZ14a}, which generally contain sequence of well know nonlinear operations like convolution, pooling, and single non-linearities like rectified linear units activations  \cite{icml2010_NairH10}. As it is shown in \cite{DBLP:journals/corr/TranBFTP14} that  such approaches could be generalized  to a three-dimensional case  to build representational vectors of local movements in videos.

\par
\begin{figure*}[t]
  \centering
  \includegraphics[scale=0.4]{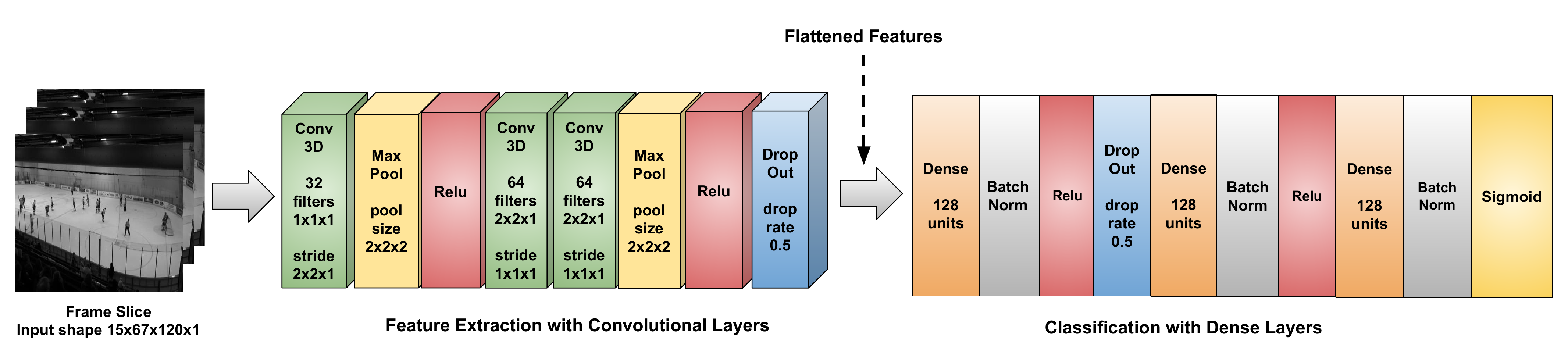}
  \caption{General Structure of the network.}
  \label{fig:graph0_h}
\end{figure*}

 The network structure is summarized in  Fig. \ref{fig:graph0_h}.
It is important to mention that this structure is chosen using a series of incremental experiments on training and validation data for the \textit{Play} class from the chosen dataset, which will be explained in the next section, in a one-against-all setting. 
During this search, we always balance between the performance and the number of parameters.  
Once found and validated for the \textit{Play} class, we fix this structure as the basis for all other cases. It should also be noted that for training our models, we use binary cross-entropy as the loss function.
%

\section{Experiments and Results}

\subsection{Dataset}
The dataset that is used in this work was presented in \cite{DBLP:conf/iscas/CarbonneauRGG15}. The paper presented a two-staged hierarchical method, based on classical computer vision, for play-break detection in non-edited hockey videos. The dataset consists of 36 gray scale videos having a $480 \times 270$ pixels resolution captured at 30 frames per second. Alls videos were recorded using a static camera. 
%

There are  12 types of events in this dataset. Full list of events. Detailed explanation can be found in the original paper \cite{DBLP:conf/iscas/CarbonneauRGG15}.
Every frame of a video is labeled with a binary string. For example, a frame having a label of 00000000101 means that this frame is associated with classes \textit{Shot} and \textit{Play}.


\subsection{Metrics}
\par We choose $F_{1}$ score as the evaluation metric for our experiments, since $F_{1}$ score is one of the recommended metrics to be used in the case of unbalanced data \cite{DBLP:journals/tkde/ZhangZ14}. We can define $F_{1}$ score as:
\begin{equation}
F_{1} = \frac{2PR} {P+R}
\end{equation}
where
$R =  \frac{T_{p}}{T_{p} + F_{n}}$ is the recall,
$P = \frac{T_{p}}{T_{p} + F_{p}}$ is the precision, $T_{p}$ mean true positives, $F_{p}$ means false positives, and $F_{n}$ means false negatives.

\subsection{Experiments}
To evaluate our work, we perform a series of experiments, which is as follows.
\begin{table}[t]
  \centering
  \caption{$F_{1}$ scores for every action. Ensemble model (EM), Single Multi-label $k$-output Model (SMKO)} 
  \label{table:full}
  \begin{tabular}{|c|c|c|}
  \hline 
  Event & EM & SMKO  \\
  \hline
  Celebration & 0.60 & 0.62 \\
  \hline
  Checking & 0.20 & \textbf{0.38} \\
  \hline
  End of period & 0.90 & \textbf{0.98}\\
  \hline
  Fight & 0.74 & \textbf{0.85} \\
  \hline
  Goal & 0.38 & 0.38  \\
  \hline
  Penalty & 0.54 & \textbf{0.79} \\
  \hline
  Shot & \textbf{0.46} & 0.30 \\
  \hline
  Save & \textbf{0.60} & 0.46 \\
  \hline
  Line change & 0.68 & \textbf{0.78}\\
  \hline
  Face off & 0.78 & \textbf{0.86} \\
  \hline
  Play & 0.93 & \textbf{0.95} \\
  \hline
  Average $F_{1}$ & 0.62 & \textbf{0.67}\\
  \hline 
  \end{tabular}
\end{table}

\subsubsection{Ensemble Model vs. Single Multi-label $k$-output Model}
The purpose of this experiment is to understand which model works better. For this, we take the basic structure, which we have described previously, and apply it to both strategies. The results of this experiment for all activities are summarized in second and third columns of Table \ref{table:full}, respectively.


\subsubsection{Comparison with Original work} 
The purpose of this experiment is to compare our deep learning approach for action recognition of the hockey game with the original work on the chosen dataset \cite{DBLP:conf/iscas/CarbonneauRGG15}. 
The results of this experiment for the three activities (as was done in \cite{DBLP:conf/iscas/CarbonneauRGG15}) are summarized in Table \ref{table:comp}.

\begin{table}[ht]
  \centering
  \caption{Comparison with original work \cite{DBLP:conf/iscas/CarbonneauRGG15} in terms of $F_{1}$ score. Ensemble Model (EM), Single Multi-label $k$-output Model (SMKO) }
  \label{table:comp}
  \begin{tabular}{|c|c|c|c|}
  \hline 
  Event & Original paper & EM & SMKO \\
  \hline
  Line change & 0.52 & 0.68 & \textbf{0.78}\\
  \hline
  Face-off & 0.36 & 0.78 & \textbf{0.86}\\
  \hline
  Play & 0.86 & 0.93 & \textbf{0.95}\\
  \hline 
  \end{tabular}
\end{table}

\begin{table}[t]
 	\centering
    \caption{$F_{1}$ scores for every action category for single multi-label $k$-output model ($F_{1}$ score of 0.67) after removing: (D) weights adjustment, (B) data normalization, (C) threshold softening, (D) all of the previous steps}
  \label{table:eval}
    \begin{tabular}{|c|c|c|c|c|}
    \hline
    \textit{Event}  & (A) & (B) & (C) & (D)  \\ 
    \hline
    Celebration     & 0.0    & 0.35   & 0.26   & 0.0  \\ 
    \hline
    Checking  & 0.1    & 0.16   & 0.09   & 0.0  \\ 
    \hline
    End of period   & 0.83   & 0.87   & 0.93   & 0.66 \\ 
    \hline
    Fight     & 0.48   & 0.43   & 0.83   & 0.0  \\ 
    \hline
    Goal & 0.003  & 0.22   & 0.0    & 0.0  \\ 
    \hline
    Penalty   & 0.6    & 0.33   & 0.63   & 0.0  \\ 
    \hline
    Shot & 0.12   & 0.07   & 0.03   & 0.0  \\ 
    \hline
    Save & 0.23   & 0.16   & 0.14   & 0.0  \\ 
    \hline
    Line change & 0.64   & 0.51   & 0.74   & 0.07 \\ 
    \hline
    Face off  & 0.73   & 0.62   & 0.79   & 0.27 \\ 
    \hline
    Play & 0.91   & 0.85   & 0.95   & 0.81 \\ 
    \hline
    Average $F_{1}$ & 0.42   & 0.42   & 0.49   & 0.17 \\ 
    \hline
    \end{tabular}
\end{table}

 \par \textit{Evaluating the Use of Data Normalization}.The purpose of this experiment is to evaluate our claim data normalization is important for achieving high recognition accuracy. To perform this experiment, we use only the single multi-label $k$-output model, as it provided the best results in the previous experiments. We remove the data normalization component and repeat the same settings as in the first experiment. The results for all activities are summarized in column (B) of Table \ref{table:eval}.

 \par \textit{Evaluating the Use of Class Weights Adjustment}.The purpose of this experiment is to evaluate our claim that handling the data unbalancing problem is important for achieving high recognition accuracy. We repeat the same settings as in the fourth experiment, but this time we do the data preprocessing and remove the class weights adjustment part instead. The results for all activities are summarized in column (A) of Table \ref{table:eval}.

\par \textit{Evaluating the Use Threshold Softening}. The purpose of this experiment is to evaluate our claim that threshold softening is important for achieving high recognition accuracy. We repeat the experiment, keeping everything except the threshold softening part of the system and use a constant threshold, instead. The results are summarized in column (C) of Table \ref{table:eval}. 

 \par \textit{Evaluating the Use of All}. The purpose of this experiment is to study how the system would perform if we removed all of the above components. We repeat the experiment, but we do not perform any data normalization, weights adjustment as well as threshold softening. The results are summarized in column (D) of Table \ref{table:eval}.

\section{Discussion}
\begin{figure}[t]
  \centering
  \subfigure{\includegraphics[scale=0.4]{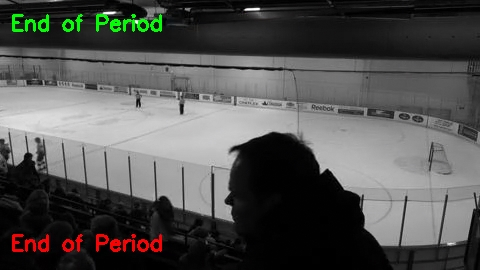} 
  \hfill \includegraphics[scale=0.4]{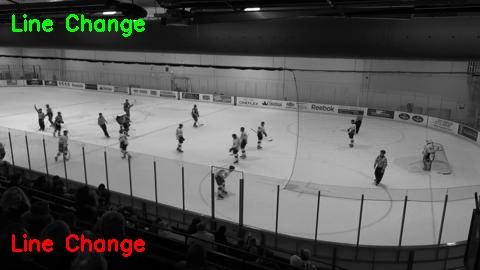}}

  \subfigure{\includegraphics[scale=0.4]{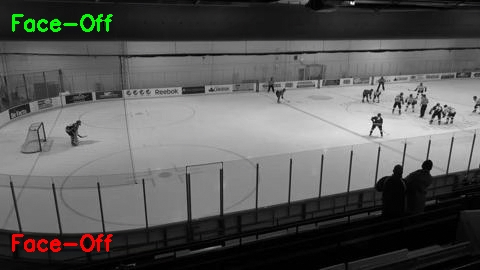}
  \hfill \includegraphics[scale=0.4]{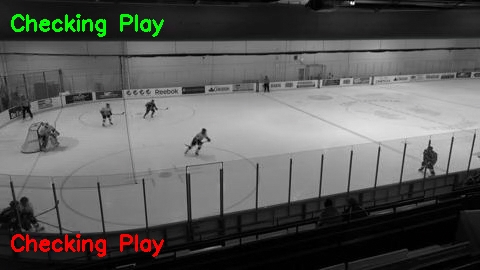}}

  \caption{Real-time Demo: Green labels are the ground truth, Red labels are the predictions}
  \label{fig:demo}
\end{figure}

Fig. \ref{fig:demo} shows visual real-time performance of our system in four situations: \textit{end of a period, line change, face-off,} and  \textit{checking-play} (multi-label case).

\par Furthermore, based on the results of ensemble model versus the single model, we can see that the single model approach is better ($F_{1}$ score of 0.62 vs. 0.67 on average, respectively). Another natural advantage of using the single model is that it can be trained $k$-times faster since it has the same number of weights as one model in the ensemble of $k$-networks (980,000). 


\par If we take a look at the results of the experiments that evaluate the importance of data normalization, class weights adjustment and threshold softening, we may see that when applied all together these steps increase the average $F_{1}$ performance significantly compared to the case where we use none of them (0.67 vs. 0.17). Furthermore, solving the class imbalance problem seems to have the most positive influence on the most under-represented classes, such as \textit{Checking}; see its associated rows in Table \ref{table:full} column SMKO and Table \ref{table:eval} columns (A) and (C). On the other hand pre-processing seems to be important for almost all classes, e.g., \textit{Play}; see its associated row in Table \ref{table:eval} columns (A) and (B). As for the threshold calibration, it is also important for all cases. In some cases, such as \textit{Celebration}, it can increase individual performance of an action by up-to 35\%. However we need to pay further attention to improving these procedures, such as implementing new algorithms for balancing in multi-label case or estimation of calibration thresholds in a different way, e.g., by learning or maximizing some criteria.
\par According to the obtained results, presented in Table \ref{table:full}, it can be seen that the system performed well for actions like \textit{Play}, \textit{Face off}, and \textit{End of Period}, with scores of 0.78 - 0.95, However, for other actions, such as \textit{Shot}, \textit{Save}, and \textit{Celebration}, the performance is not in the same range, having scores of 0.38 - 0.62.

\par We think that we got good results for \textit{Play}, \textit{Face off},  and \textit{End of Period} because we can consider these actions as group actions - interaction among players. Thus we may conclude that for group situations, our system is capable of learning adequate features to achieve an optimum recognition accuracy. Although  \textit{Celebration} can also be classified as a group action; it has its own specific challenges. For humans, primary signal of celebration is raising of the hockey sticks by players. However, the same phenomena may confuse the model. 


\par As for the \textit{Save}, \textit{Shot}, \textit{Goal} and \textit{Checking}, one probable reason for low recognition accuracy could be the fact that these events are related to the movements of specific players. 
Also, for \textit{Shot} and \textit{Save}, the ensemble model has shown better performance than the single model, but for other events, we have opposite results. 

\par It is important to mention the reason behind using the soft-thresholding methodology in the case of single $k$-output model and why it works. As it is mentioned earlier, we cannot use oversampling for the ensemble-model as it may affect the class distribution under multi-label settings. Therefore, to favor the minority classes, we used class weights adjustment and assigned higher weights to the minority classes. When we tested our system with only class-weights implementation, we got significantly low $F_{1}$ scores. Since we have unbalanced data, where positive classes are a minority, the reason for having a low $F_{1}$ score was the low recall. One way to avoid this is by increasing the complexity of the model, but this would have gone against our requirement of the real-time working of the system. Therefore, to make the model favor minority classes and penalize majority classes, besides using higher weights for the former we also use softer thresholds for them, which increases the recall scores resulting in a better $F_{1}$ score. 



\par Finally, another very important question is to understand how different deep architectures, especially non-VGG like, will work for the chosen problem. It may be reasonable to try to use different combinations and variations of CNN-LSTM\cite{DBLP:conf/iccv/SunJCYSS17}\cite{DBLP:journals/spl/WangGSS17} to try to use multi-input networks to provide more information and learn better representations. 

\section{Conclusion}
In this paper, we present deep learning based solution for hockey game action recognition in multi-label learning settings having class imbalance problem. The proposed system achieved good performance for several action categories, and it can be adapted for real-time use, although this might require the use of a specific hardware. 
As a part of our contributions, we present baseline $F_{1}$ scores for all action categories in a publicly available hockey videos dataset. Our results are better than the existing solution, and it can be a starting point for further research using this dataset.

\bibliographystyle{IEEEtran}
\bibliography{IEEEabrv,bibliography}

\begin{thebibliography}{10}
\providecommand{\url}[1]{#1}
\csname url@samestyle\endcsname
\providecommand{\newblock}{\relax}
\providecommand{\bibinfo}[2]{#2}
\providecommand{\BIBentrySTDinterwordspacing}{\spaceskip=0pt\relax}
\providecommand{\BIBentryALTinterwordstretchfactor}{4}
\providecommand{\BIBentryALTinterwordspacing}{\spaceskip=\fontdimen2\font plus
\BIBentryALTinterwordstretchfactor\fontdimen3\font minus
  \fontdimen4\font\relax}
\providecommand{\BIBforeignlanguage}[2]{{%
\expandafter\ifx\csname l@#1\endcsname\relax
\typeout{** WARNING: IEEEtran.bst: No hyphenation pattern has been}%
\typeout{** loaded for the language `#1'. Using the pattern for}%
\typeout{** the default language instead.}%
\else
\language=\csname l@#1\endcsname
\fi
#2}}
\providecommand{\BIBdecl}{\relax}
\BIBdecl

\bibitem{Zhu:2009:HDA:1703775.1704035}
\BIBentryALTinterwordspacing
C.~Zhu and W.~Sheng, ``Human daily activity recognition in robot-assisted
  living using multi-sensor fusion,'' in \emph{Proceedings of the 2009 IEEE
  International Conference on Robotics and Automation}, ser. ICRA'09.\hskip 1em
  plus 0.5em minus 0.4em\relax Piscataway, NJ, USA: IEEE Press, 2009, pp.
  3644--3649. [Online]. Available:
  \url{http://dl.acm.org/citation.cfm?id=1703775.1704035}
\BIBentrySTDinterwordspacing

\bibitem{DBLP:journals/mbec/KhanLLK10}
\BIBentryALTinterwordspacing
A.~M. Khan, Y.~Lee, S.~Lee, and T.~Kim, ``Accelerometer's position independent
  physical activity recognition system for long-term activity monitoring in the
  elderly,'' \emph{Med. Biol. Engineering and Computing}, vol.~48, no.~12, pp.
  1271--1279, 2010. [Online]. Available:
  \url{https://doi.org/10.1007/s11517-010-0701-3}
\BIBentrySTDinterwordspacing

\bibitem{Lee_EMBC_2010}
M.~W. Lee, A.~M. Khan, J.~H. Kim, Y.~S. Cho, and T.~S. Kim, ``A single
  tri-axial accelerometer-based real-time personal life log system capable of
  activity classification and exercise information generation,'' in \emph{2010
  Annual International Conference of the IEEE Engineering in Medicine and
  Biology}, Aug 2010, pp. 1390--1393.

\bibitem{DBLP:journals/tmm/HuangSC06}
\BIBentryALTinterwordspacing
C.~Huang, H.~Shih, and C.~Chao, ``Semantic analysis of soccer video using
  dynamic bayesian network,'' \emph{{IEEE} Trans. Multimedia}, vol.~8, no.~4,
  pp. 749--760, 2006. [Online]. Available:
  \url{https://doi.org/10.1109/TMM.2006.876289}
\BIBentrySTDinterwordspacing

\bibitem{Johansson1973}
G.~Johansson, ``Visual perception of biological motion and a model for its
  analysis,'' \emph{Perception \& psychophysics}, vol.~14, no.~2, pp. 201--211,
  1973.

\bibitem{Aggarwal:2011:HAA:1922649.1922653}
\BIBentryALTinterwordspacing
J.~Aggarwal and M.~Ryoo, ``Human activity analysis: A review,'' \emph{ACM
  Comput. Surv.}, vol.~43, no.~3, pp. 16:1--16:43, Apr. 2011. [Online].
  Available: \url{http://doi.acm.org/10.1145/1922649.1922653}
\BIBentrySTDinterwordspacing

\bibitem{Bulling:2014:THA:2578702.2499621}
\BIBentryALTinterwordspacing
A.~Bulling, U.~Blanke, and B.~Schiele, ``A tutorial on human activity
  recognition using body-worn inertial sensors,'' \emph{ACM Comput. Surv.},
  vol.~46, no.~3, pp. 33:1--33:33, Jan. 2014. [Online]. Available:
  \url{http://doi.acm.org/10.1145/2499621}
\BIBentrySTDinterwordspacing

\bibitem{Saputri_IJDSN_2014}
\BIBentryALTinterwordspacing
T.~R.~D. Saputri, A.~M. Khan, and S.-W. Lee, ``User-independent activity
  recognition via three-stage ga-based feature selection,'' \emph{International
  Journal of Distributed Sensor Networks}, vol.~10, no.~3, p. 706287, 2014.
  [Online]. Available: \url{https://doi.org/10.1155/2014/706287}
\BIBentrySTDinterwordspacing

\bibitem{He:2009:LID:1591901.1592322}
\BIBentryALTinterwordspacing
H.~He and E.~A. Garcia, ``Learning from imbalanced data,'' \emph{IEEE Trans. on
  Knowl. and Data Eng.}, vol.~21, no.~9, pp. 1263--1284, Sep. 2009. [Online].
  Available: \url{http://dx.doi.org/10.1109/TKDE.2008.239}
\BIBentrySTDinterwordspacing

\bibitem{DBLP:journals/tkde/ZhangZ14}
\BIBentryALTinterwordspacing
M.~Zhang and Z.~Zhou, ``A review on multi-label learning algorithms,''
  \emph{{IEEE} Trans. Knowl. Data Eng.}, vol.~26, no.~8, pp. 1819--1837, 2014.
  [Online]. Available: \url{http://dx.doi.org/10.1109/TKDE.2013.39}
\BIBentrySTDinterwordspacing

\bibitem{DBLP:journals/corr/SimonyanZ14a}
\BIBentryALTinterwordspacing
K.~Simonyan and A.~Zisserman, ``Very deep convolutional networks for
  large-scale image recognition,'' \emph{CoRR}, vol. abs/1409.1556, 2014.
  [Online]. Available: \url{http://arxiv.org/abs/1409.1556}
\BIBentrySTDinterwordspacing

\bibitem{DBLP:journals/corr/RenHG015}
\BIBentryALTinterwordspacing
S.~Ren, K.~He, R.~B. Girshick, and J.~Sun, ``Faster {R-CNN:} towards real-time
  object detection with region proposal networks,'' \emph{CoRR}, vol.
  abs/1506.01497, 2015. [Online]. Available:
  \url{http://arxiv.org/abs/1506.01497}
\BIBentrySTDinterwordspacing

\bibitem{DBLP:journals/corr/ValmadreBHVT17}
\BIBentryALTinterwordspacing
J.~Valmadre, L.~Bertinetto, J.~F. Henriques, A.~Vedaldi, and P.~H.~S. Torr,
  ``End-to-end representation learning for correlation filter based tracking,''
  \emph{CoRR}, vol. abs/1704.06036, 2017. [Online]. Available:
  \url{http://arxiv.org/abs/1704.06036}
\BIBentrySTDinterwordspacing

\bibitem{Goodfellowbook}
\BIBentryALTinterwordspacing
{Ian Goodfellow, Yoshua Bengio, and Aaron Courville}, ``Deep learning,'' 2016,
  book in preparation for MIT Press. [Online]. Available:
  \url{http://www.deeplearningbook.org}
\BIBentrySTDinterwordspacing

\bibitem{DBLP:conf/iscas/CarbonneauRGG15}
\BIBentryALTinterwordspacing
M.~Carbonneau, A.~J. Raymond, E.~Granger, and G.~Gagnon, ``Real-time visual
  play-break detection in sport events using a context descriptor,'' in
  \emph{2015 {IEEE} International Symposium on Circuits and Systems, {ISCAS}
  2015, Lisbon, Portugal, May 24-27, 2015}, 2015, pp. 2808--2811. [Online].
  Available: \url{http://dx.doi.org/10.1109/ISCAS.2015.7169270}
\BIBentrySTDinterwordspacing

\bibitem{DBLP:journals/corr/WaltnerMB14}
\BIBentryALTinterwordspacing
G.~Waltner, T.~Mauthner, and H.~Bischof, ``Indoor activity detection and
  recognition for sport games analysis,'' \emph{CoRR}, vol. abs/1404.6413,
  2014. [Online]. Available: \url{http://arxiv.org/abs/1404.6413}
\BIBentrySTDinterwordspacing

\bibitem{Tora_2017_CVPR_Workshops}
M.~Roy~Tora, J.~Chen, and J.~J. Little, ``Classification of puck possession
  events in ice hockey,'' in \emph{The IEEE Conference on Computer Vision and
  Pattern Recognition (CVPR) Workshops}, July 2017.

\bibitem{BMVC.23.124}
H.~Wang, M.~M. Ullah, A.~Klaser, I.~Laptev, and C.~Schmid, ``Evaluation of
  local spatio-temporal features for action recognition,'' in \emph{Proceedings
  of the British Machine Vision Conference}.\hskip 1em plus 0.5em minus
  0.4em\relax BMVA Press, 2009, pp. 124.1--124.11, doi:10.5244/C.23.124.

\bibitem{NIPS2012_4824}
\BIBentryALTinterwordspacing
A.~Krizhevsky, I.~Sutskever, and G.~E. Hinton, ``Imagenet classification with
  deep convolutional neural networks,'' in \emph{Advances in Neural Information
  Processing Systems 25}, F.~Pereira, C.~J.~C. Burges, L.~Bottou, and K.~Q.
  Weinberger, Eds.\hskip 1em plus 0.5em minus 0.4em\relax Curran Associates,
  Inc., 2012, pp. 1097--1105. [Online]. Available:
  \url{http://papers.nips.cc/paper/4824-imagenet-classification-with-deep-convolutional-neural-networks.pdf}
\BIBentrySTDinterwordspacing

\bibitem{DBLP:journals/corr/IbrahimMDVM15}
\BIBentryALTinterwordspacing
M.~Ibrahim, S.~Muralidharan, Z.~Deng, A.~Vahdat, and G.~Mori, ``A hierarchical
  deep temporal model for group activity recognition,'' \emph{CoRR}, vol.
  abs/1511.06040, 2015. [Online]. Available:
  \url{http://arxiv.org/abs/1511.06040}
\BIBentrySTDinterwordspacing

\bibitem{Karpathy:2014:LVC:2679600.2680211}
\BIBentryALTinterwordspacing
A.~Karpathy, G.~Toderici, S.~Shetty, T.~Leung, R.~Sukthankar, and L.~Fei-Fei,
  ``Large-scale video classification with convolutional neural networks,'' in
  \emph{Proceedings of the 2014 IEEE Conference on Computer Vision and Pattern
  Recognition}, ser. CVPR '14.\hskip 1em plus 0.5em minus 0.4em\relax
  Washington, DC, USA: IEEE Computer Society, 2014, pp. 1725--1732. [Online].
  Available: \url{http://dx.doi.org/10.1109/CVPR.2014.223}
\BIBentrySTDinterwordspacing

\bibitem{DBLP:journals/corr/KayCSZHVVGBNSZ17}
\BIBentryALTinterwordspacing
W.~Kay, J.~Carreira, K.~Simonyan, B.~Zhang, C.~Hillier, S.~Vijayanarasimhan,
  F.~Viola, T.~Green, T.~Back, P.~Natsev, M.~Suleyman, and A.~Zisserman, ``The
  kinetics human action video dataset,'' \emph{CoRR}, vol. abs/1705.06950,
  2017. [Online]. Available: \url{http://arxiv.org/abs/1705.06950}
\BIBentrySTDinterwordspacing

\bibitem{DBLP:books/sp/datamining2005/Chawla05}
N.~V. Chawla, ``Data mining for imbalanced datasets: An overview,'' in
  \emph{The Data Mining and Knowledge Discovery Handbook.}, 2005, pp. 853--867.

\bibitem{icml2010_NairH10}
\BIBentryALTinterwordspacing
V.~Nair and G.~E. Hinton, ``Rectified linear units improve restricted boltzmann
  machines,'' in \emph{Proceedings of the 27th International Conference on
  Machine Learning (ICML-10)}, J.~Fürnkranz and T.~Joachims, Eds.\hskip 1em
  plus 0.5em minus 0.4em\relax Omnipress, 2010, pp. 807--814. [Online].
  Available: \url{http://www.icml2010.org/papers/432.pdf}
\BIBentrySTDinterwordspacing

\bibitem{DBLP:journals/corr/TranBFTP14}
\BIBentryALTinterwordspacing
D.~Tran, L.~D. Bourdev, R.~Fergus, L.~Torresani, and M.~Paluri, ``{C3D:}
  generic features for video analysis,'' \emph{CoRR}, vol. abs/1412.0767, 2014.
  [Online]. Available: \url{http://arxiv.org/abs/1412.0767}
\BIBentrySTDinterwordspacing

\bibitem{DBLP:conf/iccv/SunJCYSS17}
\BIBentryALTinterwordspacing
L.~Sun, K.~Jia, K.~Chen, D.~Yeung, B.~E. Shi, and S.~Savarese, ``Lattice long
  short-term memory for human action recognition,'' in \emph{{IEEE}
  International Conference on Computer Vision, {ICCV} 2017, Venice, Italy,
  October 22-29, 2017}, 2017, pp. 2166--2175. [Online]. Available:
  \url{https://doi.org/10.1109/ICCV.2017.236}
\BIBentrySTDinterwordspacing

\bibitem{DBLP:journals/spl/WangGSS17}
\BIBentryALTinterwordspacing
X.~Wang, L.~Gao, J.~Song, and H.~T. Shen, ``Beyond frame-level {CNN:}
  saliency-aware 3-d {CNN} with {LSTM} for video action recognition,''
  \emph{{IEEE} Signal Process. Lett.}, vol.~24, no.~4, pp. 510--514, 2017.
  [Online]. Available: \url{https://doi.org/10.1109/LSP.2016.2611485}
\BIBentrySTDinterwordspacing

\end{thebibliography}

\end{document}